\documentclass{article}
\usepackage{spconf,amsmath,graphicx,hyperref}
\usepackage{booktabs} 
\usepackage{cite}
\usepackage{amssymb,amsfonts}
\usepackage{algorithmic}
\usepackage{graphicx}
\usepackage{textcomp}
\usepackage{xcolor}
\usepackage{twemojis}
\usepackage[dvipsnames]{xcolor}
\usepackage{makecell}
\usepackage[table]{xcolor}
\usepackage{booktabs}
\usepackage{multirow}
\usepackage{graphicx}
\usepackage[margin=1in]{geometry}
\usepackage{colortbl}
\usepackage{threeparttable}

\usepackage{listings}
\selectcolormodel{natural}
\usepackage{ninecolors}
\selectcolormodel{rgb}
\usepackage{tabularray}
\usepackage{booktabs,makecell}
\usepackage{wrapfig}
\usepackage[most]{tcolorbox}
\usepackage{enumitem}
\usepackage{twemojis}
\usepackage{multirow}
\usepackage{csquotes}
\usepackage{arydshln}
\usepackage{tikz}
\usepackage{pgfplots}
\pgfplotsset{compat=1.18}
\usepackage{subfigure}
\usepackage{fix-cm}
\usepackage{pifont}

\newcommand{\yes}{\scalebox{1}{\twemoji{check mark button}}}
\newcommand{\no}{\scalebox{1}{\twemoji{cross mark}}}
\newcommand{\mycc}{\cellcolor{Melon}}
\newcommand{\myc}{\cellcolor{LimeGreen}}
\newcommand{\myb}{\cellcolor{GreenYellow}}

\definecolor{darkyellow}{RGB}{251,188,4}
\definecolor{darkgreen}{RGB}{52,168,83}
\definecolor{lightblue}{RGB}{66,133,244}
\definecolor{acqua}{RGB}{70,189,198}
\definecolor{tuluBlue}{HTML}{2063BA}
\definecolor{tuluLightBlue}{HTML}{74B0FF}
\definecolor{tuluGray}{HTML}{333333}

\newcommand{\MLMA}{\includegraphics[width=15px]{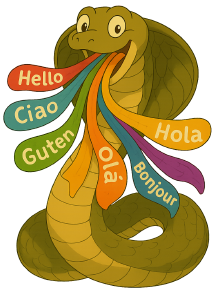}}
\title{\MLMA \hspace{0.01in} \textbf{\Large MLMA:} Towards \underline{M}ulti\underline{L}ingual ASR with \underline{MA}mba-Based Architectures }
\name{Mohamed Nabih Ali, Daniele Falavigna, Alessio Brutti\thanks{$^{*}$ We acknowledge the support of the PNRR project FAIR - Future AI
Research (PE00000013), under the NRRP MUR program funded by the
NextGenerationEU. \\
$^{**}$ We acknowledge the CINECA award under the ISCRC initiative, for the
availability of high performance computing resources and support.”}}
\address{Center for Augmented Intelligence, Fondazione Bruno Kessler, Trento, Italy}
\begin{document}
\ninept
\maketitle
\begin{abstract}
Multilingual automatic speech recognition (ASR) remains a challenging task, especially when balancing performance across high- and low-resource languages. Recent advances in sequence modeling suggest that architectures beyond Transformers may offer better scalability and efficiency. In this work, we introduce MLMA (Multilingual Language Modeling with Mamba for ASR), a new approach that leverages the Mamba architecture—an efficient state-space model optimized for long-context sequence processing—for multilingual ASR. Using Mamba, MLMA implicitly incorporates language-aware conditioning and shared representations to support robust recognition across diverse languages. Experiments on standard multilingual benchmarks show that MLMA achieves competitive performance compared to Transformer-based architectures. These results highlight Mamba’s potential as a strong backbone for scalable, efficient, and accurate multilingual speech recognition.
\end{abstract}
\begin{keywords}
Multi-lingual ASR, State Space Models, Mamba
\end{keywords}
\section{Introduction}
\label{sec:intro}

Automatic Speech Recognition (ASR) has become a cornerstone of modern computing, supporting applications such as voice assistants, transcription services, and real-time speech translation. Driven by large-scale datasets and deep learning advances, ASR systems have reached near-human performance in high-resource languages like English and Mandarin \cite{stoian2020analyzing, zhang2025speech}. However, most existing systems are language-specific, which limits scalability and exacerbates the performance gap for low-resource languages with limited annotated data \cite{qian2024learn}.

Multilingual ASR has emerged as a promising alternative by training a single model across multiple languages \cite{pratap2020massively, li2021scaling}. Such models exploit shared phonetic and acoustic representations, enabling cross-lingual transfer from high-resource to under-represented languages. Despite this potential, achieving robust multilingual performance remains challenging. Transformer-based architectures, now dominant in ASR \cite{loubser2024end, kim2022squeezeformer}, provide strong sequence modeling capabilities but with high computational and memory costs. These inefficiencies are especially problematic in multilingual scenarios, where diverse speech rates, prosodic patterns, and phenomena such as code-switching demand processing of long and variable-length utterances. 

Recently, Mamba architecture~\cite{gu2023mamba} has been proposed to handle variable-length input sequences and temporal irregularities, common in multilingual speech data. Therefore, it can generalize across languages with different rhythmic and phonetic structures. Mamba also supports streaming ASR with mechanisms like lookahead and unimodal aggregation (UMA), which help it adapt to real-time multilingual input \cite{fang2025mamba}. 
These features are particularly beneficial for languages characterized by rapid speech transitions or tonal variations, where conventional models often struggle to maintain recognition accuracy and latency.



Integrating Mamba into multilingual ASR offers several advantages: its memory-efficient design lowers training and inference costs \cite{qu2024survey}, its sequential inductive bias can better capture cross-lingual phonetic structures—benefiting code-switching and low-resource languages—and its scalability enables adding languages without proportional computational overhead. 

In this work, we investigate the application of Mamba-based architectures to multilingual ASR. \ding{182} We conducted experiments across a diverse set of European languages. \ding{183} Analyzing their ability to learn shared linguistic representations by comparing performance against Transformer-based baselines, and their robustness to multilingual challenges. Our goal is to bridge the gap between recent advances in efficient sequence modeling and the development of inclusive, scalable ASR systems that can serve a truly global user base. {\bf Our MLMA model, trained on almost 12K hours covering 6 languages, is the first multilingual ASR system based on Mamba}. MLMA code and weights are publicly available under the most permissive license.

\section{Related Works}
\label{sec:format}
With the advent of deep learning, multilingual ASR systems, capable of recognizing multiple languages, has grown in demand for cross-lingual use~\cite{li2025efficient}. Recent work in multilingual ASR has drastically increased language coverage to support hundreds and even thousands of languages. This includes approaches based on labeled training data such as Whisper \cite{radford2023robust}, USM \cite{zhang2023google}, Seamless \cite{barrault2023seamlessm4t} and MMS \cite{pratap2024scaling}, Ml-superb 2.0 \cite{shi2024ml}, FAMA \cite{papi2025fama} as well as zero-shot work \cite{zhao2025scaling}. While these transformer-based approaches are highly effective for modeling long-range dependencies, Transformers have notable drawbacks: their quadratic complexity makes long-sequence processing costly, they require vast amounts of labeled or weakly labeled data that are scarce for low-resource languages, and their large size limits their deployment in resource-constrained settings \cite{patro2025mamba}.

\subsection{Mamba for ASR}
Mamba has been applied to various speech tasks, for example, separation and enhancement \cite{jiang2025speech, jiang2025dual, chao2024investigation}, leveraging its property of linear-time complexity to model the long sequence while maintaining low computational cost. Motivated by this, numerous studies have been conducted to evaluate Mamba's performance in ASR tasks. Table \ref{tab:related} summarizes the most recent research papers leveraging Mamba for ASR tasks, highlighting the main contribution of each work.




\begin{table*}
    \scriptsize
    \centering
        \caption{The table summaries recent works exploring Mamba for ASR.}
    \begin{tabular}{c c c c l}
    \toprule
         \bf Ref. & \bf Dataset (hours) & \bf  Multilingual & \bf Language & \bf Contribution  \\
         \midrule
         \cite{jiang2025speech}& LibriSpeech & \no & EN &  ConMamba for monolingual ASR\\ 
          \cite{fang2025mamba} & AISHELL-1\&2 & \no & Mandarin & Efficiency of Mamba for streaming ASR  \\ 
         \cite{zhang2025mamba}& LibriSpeech, AN4, SEAME, ASRU & \no & EN \& EN-Mandarin & BiMamba \\ 
         \cite{shakhadri2025samba} & LibriSpeech, GigaSpeech, SPGISpeech  & \no & EN & Samba-ASR: Mamba as Encoder and Decoder \\ 
         \cite{zhang2025rethinking} & LibriSpeech-100 & \no & EN & Mamba-based HuBERT model for ASR, \\ 
         \cite{miyazaki2024exploring} & \makecell{LibriSpeech, GigaSpeech, TEDLIUM2, \\ AISHELL, CSJ, VoxVorge}  & \no & \makecell{EN, Mandarin, \\ Japanese, IT} & Mamba performance against Transformer architectures \\ 
         \cite{lin2025exploration} & TEDLIUM3 & \no & EN & Mamba-based HuBERT  against Transformer-based SSL \\ 
         \cite{gao2024speech} & LibriSpeech-100 & \no & EN & Mamba for long-context ASR \\
         \cite{hou2024conmamba} & LibriSpeech-100 & \no & EN & Augmented ConMamba Encoder \\
        \midrule
        MLMA (ours) & \makecell{LibriSpeech, CommonVoice, \\Voxpopuli, \\ Multilingual LibriSpeech} & \yes & \makecell{EN, IT, FR, \\ ES, DE, NL} & MLMA: a European Multilingual ASR based on Mamba  \\
         \bottomrule
    \end{tabular}
    \label{tab:related}
\end{table*}

Based on the literature review in Table \ref{tab:related}, current Mamba-based ASR research exhibits significant limitations. Existing studies mainly investigate architectural replacements within Transformer backbones, but are restricted to monolingual or at most bilingual settings on small datasets like LibriSpeech-100~\cite{panayotov2015librispeech}. These works operate under matched conditions and lack the scale to test Mamba’s multilingual effectiveness. In contrast,  our proposed MLMA model explores Mamba in a  large-scale multilingual setting with nearly 12,000 hours of training across  six European languages, representing the  first multilingual ASR system based on Mamba and demonstrating its viability beyond constrained setups.

\section{Proposed Mamba architecture for multilingual ASR}
\subsection{Overview of Mamba}
Mamba is a Structured State Space Model (SSM) defined in discrete time as:
{\small
\begin{equation}
    h_{t} = \bar{A}h_{t-1} + \bar{B}x_{t},  \quad  y_{t}=Ch_{t}
 \end{equation}}
where $h_{t}$ is the state, $\bar{A}$ the transition matrix, $\bar{B}$ the input-state interaction, and $C$ the output map.

Since $\bar{A}$ and $\bar{B}$ derive from continuous-time parameters, they are not learned directly but approximated via Zero-Order Hold (ZOH):
 {\small
 \begin{equation}
     \bar{A}=\exp(\Delta A), \quad \bar{B}=(\Delta A)^{-1}(\exp(\Delta A) - I) \cdot \Delta B
 \end{equation}}
with $A, B$ the continuous forms and $\Delta$ the discretization step. Training is done on $A, B$, which are converted to $\bar{A}, \bar{B}$ at each forward pass, enabling efficient discrete-time modeling while preserving long-range dependencies. ZOH ensures that the temporal structure of the continuous-time model is retained after discretization, allowing it to track dependencies across long sequences. To increase adaptability, \cite{gu2021efficiently} introduced a selection mechanism:
 {\small
\begin{equation}
     B=f_{B}(x), \quad C=f_{C}(x), \quad \Delta= \text{Broadcast}_{D}(f_{\Delta}(x))
\end{equation}}
that is, instead of using fixed matrices $B,C,\Delta$, the model learns functions $f_B, f_C, f_{\Delta}$, that generate these parameters based on the input $x$, allowing the model to flexibly adapt its state transition and output mapping according to the current input.

Mamba \cite{gu2023mamba} extends this idea by removing the Linear Time Invariance (LTI) constraint, allowing parameters to vary over time. This improves flexibility in non-stationary environments and strengthens modeling of long-range, context-dependent behaviors.

\subsection{Convolutional Mamba (ConMamba) Encoder}
For ASR, and speech processing in general, extracting both local and global features is crucial. Models such as Conformer \cite{gulati2020conformer} and Zipformer \cite{yao2023zipformer} achieve this by combining convolution (local) with self-attention (global).  

The ConMamba Encoder follows the same principle but replaces multi-head self-attention with Mamba layers, while retaining convolution to strengthen local feature extraction. For a generic input $x$, a ConMamba encoder produces output embeddings $y$ as:
{\footnotesize
\begin{equation}
\begin{aligned}
    \tilde{x} &= x + \tfrac{1}{2} \mathrm{FFN}(x) 
    &\quad x' &= \tilde{x} + \mathrm{Mamba}(\tilde{x}) \\
    x'' &= x' + \mathrm{Conv}(x') 
    &\quad y &= \mathrm{Layernorm}\!\left(x'' + \tfrac{1}{2} \mathrm{FFN}(x'')\right)
\end{aligned}
\end{equation}
}
where {\tt FFN} is a feed-forward module, and the convolutional module which extracts local patterns. Note that the outputs of both {\tt Mamba} and {\tt Conv} layers are summed  before layer normalization with half of the output of another FFN. This hybrid design enables effective integration of local and global features for speech representation.

\subsection{Proposed Architecture}

\begin{figure}[!ht]
    \centering
    \includegraphics[width=\linewidth]{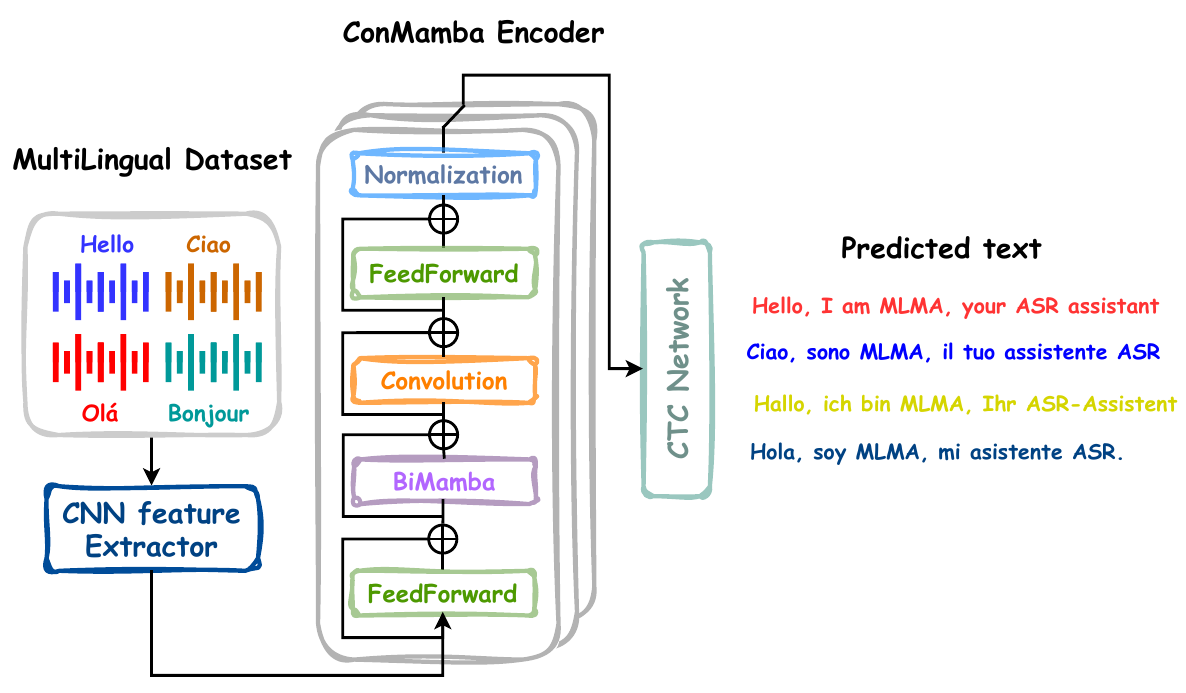}
    \caption{The architecture of MLMA using ConMamba encoder and a CTC decoder as in \cite{jiang2025speech}}
    \label{fig:MLMA arc}
\end{figure}

The proposed MLMA model, as depicted in Figure \ref{fig:MLMA arc}, follows the architecture introduced in \cite{jiang2025speech}  that integrates a convolutional transformer with a bidirectional Mamba module (Bi-Mamba) within a CTC framework. Input audio is converted to 80-dimensional log Mel filter banks, normalized, and processed by a two-block CNN for low-level feature extraction and temporal downsampling. An 18-layer Transformer encoder (hidden size 256, feed-forward 1024, dropout 0.1, GELU) models contextual representations, augmented with a Bi-Mamba module (dstate=16, expand=2, dconv=4) to capture long-range dependencies. A linear projection followed by LogSoftmax maps encoder outputs to the vocabulary (including blank, BOS, and EOS), and training is performed with CTC loss. Note that in our experiments we use the same hyperparamters reported in \footnote{\url{https://github.com/xi-j/Mamba-ASR }}. More details on the training hyperparameters, along with our implementation, is available in the public repository\footnote{\url{https://github.com/mnabihali/MLMA}}.

\section{Experimental Setup}
Our experiments leverage four large-scale multilingual speech corpora—LibriSpeech (clean subsets) \cite{panayotov2015librispeech}, CommonVoice v20.0 \cite{ardila2019common}, VoxPopuli-ASR \cite{wang2021voxpopuli}, MultiLingual LibriSpeech \cite{pratap2020mls}, and FLEURS \cite{conneau2023fleurs}. We consider 6 languages, spanning over 11,000 hours of labeled speech data in: English (en), Italian (it), French (fr), Spanish (es), German (de), and Dutch (nl). This collection combines read and semi-spontaneous speech, ensuring broad linguistic and acoustic diversity across the languages. The amount of training hours for each language and each dataset is summarized in Table~\ref{tab:data}.
\vspace{-1em}
\begin{table}[!h]
\centering
\scriptsize
\caption{List of training data used in our experiments.}
\begin{tabular}{lcccccc}
\toprule
\textbf{Dataset} & \multicolumn{6}{c}{\textbf{\#hours}} \\
\cmidrule{2-7}
 & en & it & fr & es & de & nl \\
\midrule
LS & 464 & \no & \no & \no & \no & \no \\
CV v20.0 & 1774 & 249 & 829 & 499 & 947 & 46 \\
VP-ASR & 522 & 78 & 206 & 152 & 264 & 46 \\
MLS & \no & 247 & 1077 & 918 & 1967 & 1554 \\
\midrule
FL & 7.5 & 9.0 & 10.3 & 8.8 & 9.0 & 7.7 \\
\midrule
Total: & 2760 & 574 & 2112 & 1569 & 3178 & 1646 \\
\bottomrule
\end{tabular}
\label{tab:data}
\end{table}

To assess the effectiveness of ConMamba, we compare its performance against a Conformer model \cite{gulati2020conformer} (with 18 encoder layers and hidden size equal to 256. More detail on Conformer training hyperparameters are reported in \footnote{\url{https://github.com/speechbrain/ speechbrain/blob/develop/recipes/LibriSpeech/}}) as well as we use some very large scale multilingual models (OWSM V3.1~\cite{peng2024owsm}, OWSM-CTC~\cite{peng-etal-2024-owsm}, FAMA~\cite{papi2025fama} and Whisper-Large-v3~\cite{radford2023robust}) as reference although the comparison is not fair due to different model and training sizes and different decoding mechanisms. We evaluate the performance in monolingual settings (en), bilingual (en, it) with also ablation studies and multilingual. For the latter we consider in-domain and out-of-domain data. 

\subsection{Monolingual comparison with Conformer}
Table \ref{tab:libri} compares the performance of ConMamba with a Conformer when they are both trained from scratch on Libri-1000. Note that the number of parameters of the models are rather similar. ConMamba consistently outperforms the Conformer baseline, achieving lower WER on both test-clean and test-other test sets. This indicates that the ConMamba architecture offers improved robustness and generalization over the standard Conformer design. 
\begin{table}[!htp]\centering
\caption{WER of ConMamba and Conformer on LibriSpeech dataset. Results are similar to what reported in \cite{jiang2025speech}}\label{tab:libri}
\scriptsize
\centering
\begin{tabular}{lccc}\toprule
\multirow{2}{*}{Model} &\multirow{2}{*}{\#Param(M)} &\multicolumn{2}{c}{WER(\%)($\downarrow$)} \\ \cmidrule(l{2pt}r{2pt}){3-4}
& & test-clean & test-other \\ \midrule
 
 Conformer & 28.8 & 4.27 & 11.29  \\ \midrule
 ConMamba  & 31.6 & \bf 4.05 & \bf 10.50 \\ 
\bottomrule
\end{tabular}
\vspace{-2em}
\end{table}
\subsection{Bilingual capabilities: Italian and English}
In Table~\ref{tab:bilingual} we report the performance on bilingual settings considering Italian and English. This experiment allows us to compare not only ConMamba and Conformer but also other large-scale multilingual models relying on published results. We observe that ConMamba maintains strong performance across both English and Italian, providing consistent improvements over Conformer and generalizing effectively to multilingual and less curated speech datasets. The table also compares with four multilingual very large-scale models for ASR models. Although obviously less performing due to a smaller size, less training data and a simplified training, MLMA is not that far from those models.

\begin{table}[hbt!]
\caption{WER(\%) ($\downarrow$) of bilingual ConMamba and Conformer, trained from scratch Italian and English data. Numbers for FAMA, OWSM v3.1 and Whisper-Large-v3 are from~\cite{papi2025fama}; for OSWM-CTC from~\cite{peng-etal-2024-owsm}. Note: besides having larger dimension and larger training sets, the large-scale models also employ autoregressive decoding methods.  "-": results not reported in the reference paper.} \label{tab:bilingual}
\centering
\renewcommand{\arraystretch}{1.1}
\setlength{\tabcolsep}{2.5pt}
\scriptsize
\begin{tabular}{lcccccc}
\toprule
\textbf{Model} & \multicolumn{3}{c}{\textbf{English}} & \multicolumn{3}{c}{\textbf{Italian}} \\
\cmidrule(lr){2-4} \cmidrule(lr){5-7}
 & \textbf{LS} & \textbf{CV} & \textbf{VP} & \textbf{CV} & \textbf{VP} & \textbf{MLS} \\
\midrule
ConMamba-CTC $^a$ & 3.6 & 18.8 & 10.7 & 11.4 & 24.8 & 13.4 \\ 
Conformer-CTC $^b$ & 4.4 & 22.3 & 11.5 & 14.3 & 23.7 & 14.3 \\
\midrule
FAMA $^c$~\cite{papi2025fama} & - & 13.8 & 8.9 & 7.3 & 15.7 & 12.6\\
OWSM v3.1 $^d$~\cite{peng2024owsm} & - & 11.9 & 8.4 & 12.5 & 24.0 & 19.3\\
OWSM-CTC $^d$~\cite{peng-etal-2024-owsm} & 2.4 & 12.1 & 8.6 & - & - & 22.1\\
\midrule
Whisper-Large-v3$^e$ & - & 11.2 & 7.1 & 6.5 & 18.8 & 8.8\\
\bottomrule
\end{tabular}
\vspace{0.1cm}
\begin{flushleft}
\scriptsize
$^a$ ConMamba-CTC: (31.6M-3334h). $^b$ Conformer-CTC: (28.8M-3334h). $^c$ FAMA: (475M-150K h). $^d$ OWSM models: (1020M-180K h). $^e$ Whisper large-v3: (1550M-5M).
\vspace{-3em}
\end{flushleft}
\end{table}



\subsection{Multilingual ASR}
Finally, in Table \ref{tab:multi} we evaluate the performance of an actual multilingual MLMA model that covers 6 languages and is trained on over 11840 hours of speech data. Overall, across the in-domain datasets, MLMA delivers consistent multilingual performance, effectively handling linguistic and acoustic variability in the training corpora. While performance naturally varies by language, the results indicate stable recognition capabilities across all languages. Importantly, evaluation on the unseen FLEURS benchmark further demonstrates that MLMA retains competitive performance under out-of-domain conditions, highlighting its robustness and supporting its potential as a strong foundation for multilingual ASR. Additionally, the reported results on the MLS dataset reveal that our MLMA model can achieve better performance compared to the OWSM-CTC foundation model.

\begin{table}[!ht]
\scriptsize
\caption{WER($\downarrow$\%) of MLMA across multilingual data. The numbers of OSWM-CTC are from~\cite{peng-etal-2024-owsm}. FL*: FLEURS is not used in training. "-": results not reported in the reference paper.}
\centering
\begin{tabular}{lcccccc}
\toprule
\textbf{Dataset} & \textbf{EN} & \textbf{IT} & \textbf{FR} & \textbf{ES} & \textbf{DE} & \textbf{NL} \\
\midrule
LS   & 7.2  & \no   & \no   & \no   & \no   & \no    \\
CV   & 23.2 & 13.0  & 15.0  & 11.2  & 12.9  & 16.8 \\
VP   & 11.5 & 24.5  & 14.8  & 12.9  & 16.1  & 21.5 \\
MLS  & \no  & 13.3  & 9.1   & 6.5   & 9.5   & 14.8 \\
\midrule
FL*  & 19.2 & 12.5  & 19.6  & 10.6  & 15.4  & 27.9 \\
\cmidrule(lr){1-7}
\mycc \textbf{Avg.} & \mycc \textbf{15.2} & \mycc \textbf{15.8} & \mycc \textbf{14.6} & \mycc  \textbf{10.3} & \mycc \textbf{13.5} & \mycc \textbf{20.3} \\
\midrule \midrule
\multicolumn{7}{c}{\myb OWSM-CTC}\\
\midrule \midrule
MLS &  - & 22.1 & 12.9 & 10.3 &11.9 & 20.4\\
\bottomrule
\end{tabular}
\label{tab:multi}
\end{table}


\subsection{Ablation studies}
We conclude the paper with an analysis of the impact of the model size and of the amount of training data on MLMA models in bilingual ASR. {\bf Model size:} Table \ref{tab:ablationsize} shows the performance on CV English and Italian, when scaling ConMamba from 31.6M to 42M parameters, highlighting that the model benefits from increased capacity without compromising efficiency. In particular, the larger model shows significant WER reduction on English, a language with rich phonetic diversity and complex prosody. This suggests that ConMamba can use additional parameters to refine its modeling of nuanced acoustic and linguistic patterns and to generalize to less curated datasets.
\begin{table}[!htp]\centering
\caption{WER(\%) ($\downarrow$) on CV scaling the size of MLMA}\label{tab:ablationsize}
\scriptsize
\centering
\begin{tabular}{lcccc}\toprule
\multirow{2}{*}{Model} &\multirow{2}{*}{\#Param(M)} &\multicolumn{2}{c}{WER(\%) ($\downarrow$)} \\ \cmidrule(l{2pt}r{2pt}){3-4} 
& &EN &IT \\ \cmidrule(l{2pt}r{2pt}){1-4}
\multirow{2}{*}{ConMamba} & 31.6 &  23.00 & 10.92 \\ \cmidrule(l{2pt}r{2pt}){2-4} 
& 42.0 & \bf 21.04 & \bf 10.42 \\
\bottomrule
\end{tabular}
\end{table}

\noindent{\bf Number of hours:} Table \ref{tab:ablationhours} reports the WER on CV Italian and English, while increasing the amount of training material from 710 hours to 3334 hours. The results show that increasing the size of the training data improves the bilingual ConMamba model for both languages. 
We observe consistent in-domain improvements, particularly on the CV and VP subsets, along with notable out-of-domain gains on the English portion of the FL benchmark. This trend is not observed for Italian, likely due to the increased unbalance between the languages.

This highlights that more data boosts both in-domain performance and out-of-domain robustness, confirming the scalability of the ConMamba architecture.
\begin{table}[!htbp]
\caption{WER(\%) ($\downarrow$) Performance of Bilingual ConMamba increasing the training material. 710 = LS (460h)\textsubscript{EN} + CV\textsubscript{IT} ; 1210= LS(960h)\textsubscript{EN} + CV\textsubscript{IT} and 3334 = LS(960)\textsubscript{EN} + (CV + VP)\textsubscript{EN\&IT} + MLS\textsubscript{IT}  } \label{tab:ablationhours}
\scriptsize
\centering
\renewcommand{\arraystretch}{1.2}
\setlength{\tabcolsep}{6pt}
\footnotesize
\begin{tabular}{lcccc cccc}
\toprule
& \multicolumn{4}{c}{ English} & \multicolumn{4}{c}{Italian}  \\
\cmidrule(lr){2-5} \cmidrule(lr){6-9}
\textbf{Hrs.} & \textbf{LS} & \textbf{CV} & \textbf{VP} & \textbf{\mycc FL} & \textbf{CV} & \textbf{VP} & \textbf{MLS} & \textbf{\myc FL}\\
\midrule

710 & 5.3 & 56.5 & 32.8 & \mycc 35.4 & 11.7 & 34.8 & 30.8 & \myc 10.2  \\

1210 & 3.9 & 47.1 & 24.5 & \mycc 26.3 & 11.9 & 34.9 & 31.8 & \myc 10.6\\

3334 & 3.6 & 18.8 & 10.7 & \mycc 14.9 & 11.4 & 24.8 & 13.4 & \myc 13.1\\
\bottomrule
\vspace{-2em}
\end{tabular}
\end{table}

\section{Conclusion}

In this work, we introduced MLMA, a multilingual ASR framework built upon the Mamba state-space architecture, enhanced with language-aware conditioning and shared representations. Through evaluations on standard multilingual benchmarks, MLMA demonstrated competitive recognition performance relative to Conformer-based models, while offering significantly faster inference. These findings underscore the potential of state-space models as efficient and scalable alternatives for multilingual ASR, particularly in scenarios involving both high and low-resource languages. MLMA represents a promising step toward practical ASR systems capable of real-time processing and broad linguistic coverage.

\bibliographystyle{IEEEbib}
\bibliography{refs}

\end{document}